\documentclass{article}

\usepackage[preprint]{neurips_2021}

\usepackage{bm}
\usepackage{amsmath}
\usepackage{graphicx}
\usepackage[utf8]{inputenc} % allow utf-8 input
\usepackage[T1]{fontenc}    % use 8-bit T1 fonts
\usepackage{hyperref}       % hyperlinks
\usepackage{url}            % simple URL typesetting
\usepackage{booktabs}       % professional-quality tables
\usepackage{amsfonts}       % blackboard math symbols
\usepackage{nicefrac}       % compact symbols for 1/2, etc.
\usepackage{microtype}      % microtypography
\usepackage{xcolor}         % colors

\title{Edge-featured Graph Neural Architecture Search}

\author{
Shaofei Cai$^{1,2}$, Liang Li$^{1}$\thanks{Corresponding author.}, Xinzhe Han$^{1,2}$, Zheng-Jun Zha$^{3}$, Qingming Huang$^{1,2,4}$ \\
$^{1}$Key Lab of Intell. Info. Process., Inst. of Comput. Tech., CAS, Beijing, China \\
$^{2}$University of Chinese Academy of Sciences, Beijing, China \\
$^{3}$University of Science and Technology of China, China, $^{4}$Peng Cheng Laboratory, Shenzhen, China \\
{\tt \small shaofei.cai@vipl.ict.ac.cn,liang.li@ict.ac.cn,hanxinzhe17@mails.ucas.ac.cn,} \\
{\tt \small zhazj@ustc.edu.cn, qmhuang@ucas.ac.cn}
}

\begin{document}

\maketitle

\begin{abstract}
Graph neural networks~(GNNs) have been successfully applied to learning representation on graphs in many relational tasks. 
Recently, researchers study neural architecture search~(NAS) to reduce the dependence of human expertise and explore better GNN architectures, but they over-emphasize entity features and ignore latent relation information concealed in the edges. 
% In this paper, we present edge-featured graph neural architecture search~(EGNAS) to automatically design optimal GNN architectures based on differentiable search strategy. 
To solve this problem, we incorporate edge features into graph search space and propose Edge-featured Graph Neural Architecture Search~(\textbf{EGNAS}) to find the optimal GNN architecture. 
Specifically, we design rich entity and edge updating operations to learn high-order representations, which convey more generic message passing mechanisms. 
Moreover, the architecture topology in our search space allows to explore complex feature dependence of both entities and edges, which can be efficiently optimized by differentiable search strategy. 
Experiments at three graph tasks on six datasets show EGNAS can search better GNNs with higher performance than current state-of-the-art human-designed and searched-based GNNs.\footnote{Codes have been provided in supplementary materials, and will be released via GitHub. } 
% In this paper, we propose Edge-featured Graph Neural Architecture Search~(\textbf{EGNAS}) that enables automatic design of more generic GNN architecture based on differentiable search strategy. 
% Specifically, we design novel edge-featured graph search space that consists of rich updating operations to learn higher order representation of entities and edges. 
% Specifically, we design a novel edge-featured graph search space that incorporates both entity and edge features and consists of separate feature learning operations, representing more generic message passing mechanisms. 
\end{abstract}
\section{Introduction}
% 第一段： 介绍NAS + GNAS
Architecture design is a critical component of successful deep learning. 
Lately, the neural architecture search~(NAS)~\cite{baker2016designing, liu2018darts, pham2018efficient, zhang2021automated, zoph2016neural} has been extensively studied to explore the optimal network architectures and lessen the manual intervention. 
Most NAS works~\cite{baker2016designing, liu2018darts, pham2018efficient, zoph2016neural} focus on searching architectures about CNN and RNN. 
Recently, graph neural networks~(GNNs) have become standard toolkits for analyzing complex graph-structure data, facilitating many relational tasks such as chemical molecular property prediction, social network analysis and recommendation. 
% 边、节点
However, only a few efforts~\cite{cai2021rethinking, zhang2021automated} conceive NAS in graph machine learning domain. 
%In many real-world applications, data are best constructed as graphs to analyze and display. 
In this paper, we study edge-featured graph neural architecture search to improve GNN's reasoning capability. 
% 改： propose -> study ； 去掉 edge-featured
% In this paper, we propose edge-featured graph neural architecture search~(EGNAS) that explicitly models deep relation features in graph search space to improve GNN's reasoning capability. 

% 第二段：CNN，RNN 通用的搜索空间，GNN无，特别之处。 GNN 核心是，层数设计 
The core of GNNs is the message passing mechanism, which learns graph representation by propagating feature information between neighboring entities. 
This helps capture the semantic context implied in the graph. 
Unlike image and sequence data, which have a grid structure, graph data lies in a non-Euclidean space~\cite{bronstein2017geometric}, leading to unique architectures and designs for graph machine learning. 
%For example, works~\cite{cai2021rethinking} redesign graph search space using message filtering and aggregating operations, which is more complex than typical search space with convolution and recurrent operations. 
%The most urgent task of graph NAS research is to design a more complete search space that covers effective message passing mechanisms. 
% 不同于之前搜索空间，。。。
% Different from typical search space, which consists of convolution and recurrent operations, graph search space is more complex and incomplete. 
Typical search space of CNN/RNN NAS focuses on improving the representation learning capability by combining convolution, pooling and activation operations. 
Differently, graph search space aims to explore effective message passing mechanisms, which is the key challenge to solve graph neural architecture search problem. 
%Designing complete graph search space is the most urgent task in solving graph neural architecture search problem. 
% CNN的搜索空间操作只是获取更高层次的特征，而GNN搜索空间探索消息传递机制！
% Designing graph search space to explore more effective message passing mechanisms is the most urgent task in solving graph neural architecture search problem. 

% 第三段：当前GNAS 介绍一下，总结GNAS如何对待节点和边。而近来edge-featured GNN 出现 意义所在（1，2句话）。
Currently graph search spaces can be divided into two categories: macro- and micro-space. 
Macro search space~\cite{gao2020graph, jiang2020graph} is constructed using popular GNNs (\emph{e.g.} GIN~\cite{xu2018powerful}, GAT~\cite{velivckovic2017graph}) as atom operations, %which is limited and is regarded as a kind of ensemble. 
which is actually a kind of GNN ensemble and represents limited message passing models. 
% 空间是狭小的，被看做是一种集成， and fails to 发现新的机制 specific to dataset
Micro search space consists of fine-grained atomic operations such as neighbor aggregation, feature filtering and combining function~\cite{cai2021rethinking, zhang2021automated}, that hold promise for discovering novel message passing mechanisms. 
The existing graph NAS methods mainly focus on learning effective entity embeddings, but ignore the latent high-order information associated with edges. 
In fact, mining the relation information concealed in the edges can guide fine-grained message passing on graph and enrich graph search space, which is beneficial to node-level, edge-level and graph-level tasks. 
% 对于点、边任务都有作用
% Designing powerful GNNs with better graph representation learning capability becomes the critical research direction in graph machine learning domain. 
% Benefitting from the rapid development of AutoML, recent researchers explored neural architecture search (NAS) to automatically find better GNNs specific to dataset, also known as graph neural architecture search (GNAS). 
% Due to the complexity of graph-structure data, there is neither a unified framework to design GNNs nor a unified search space to perform graph neural architecture search, which is different from CNNs and RNNs. 

% 第四段：4. 把edge featured 用好， 这些GNNs工作怎么做 ， 难以融入到搜索机制中 ， 之前 GNAS 解决不了， 需要设计新的空间 
Recently, increasing manually-designed GNNs~\cite{chen2021edge, jiang2019censnet, Li_2019_CVPR, yang2020nenn} try to incorporate edge features into models for better exploiting relation information. 
Specifically, they stack the building layers in a sequential architecture and iteratively learn features of entity and edge. 
Nevertheless, such plain architecture follows incremental updating rule based on current layer's representation, which is naive and sub-optimal. 
Also, these approaches adopt simple edge learning function, which is difficult to learn high-order and long-term relation features. 

\begin{figure*}[t]
  \centering
  \includegraphics[scale=0.50, trim = 0 0 0 0, clip]{ 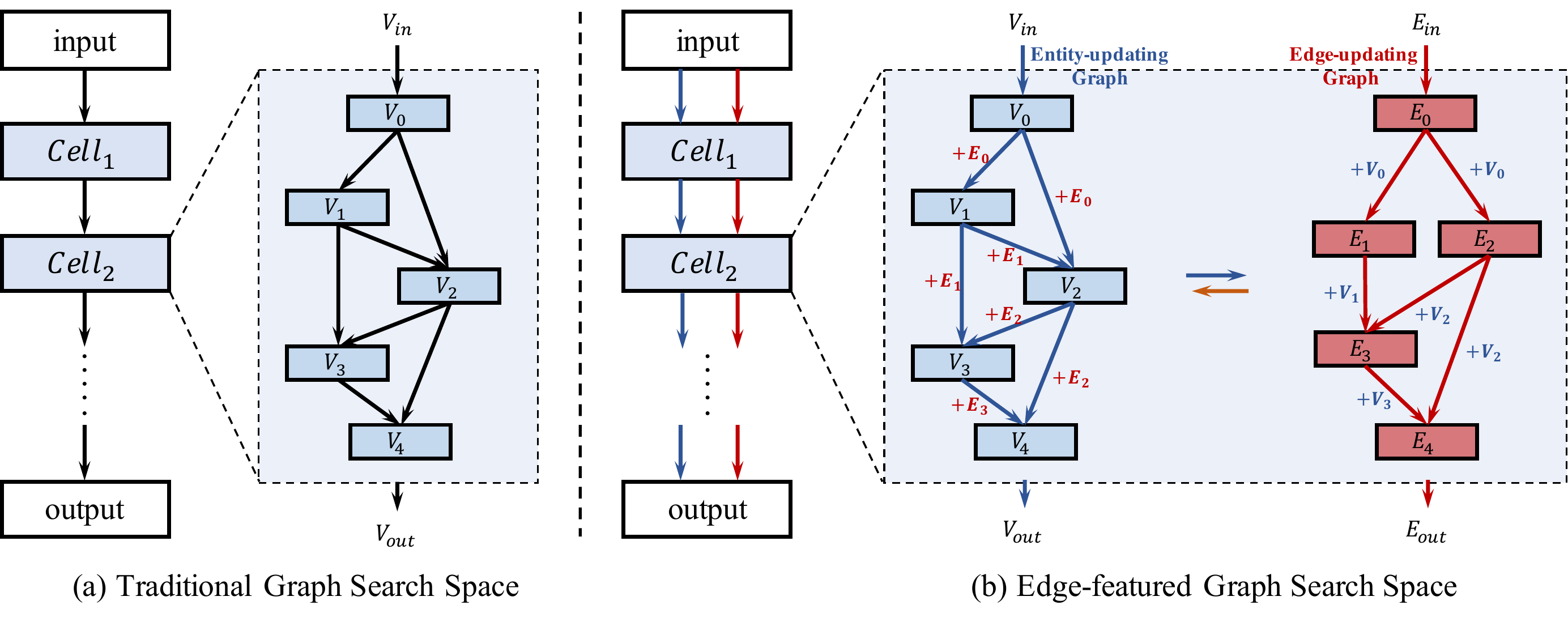}
  \caption{A comparison between the traditional cell-based graph search space and our proposed edge-featured graph search space with dual updating graph of entity and edge.}
  \label{fig_search_space}
  \vspace{-1em}
\end{figure*}

In this paper, we propose Edge-featured Graph Neural Architecture Search (EGNAS) with novel edge-featured search space to learn the optimal GNN architecture using gradient-based search strategy. 
%that explicitly models the deep relation features in search space and learns the optimal GNN architecture based on differentiable search strategy. 
% 为了捕捉不同关系之间的依赖性，我们的搜索空间允许复杂的关系依赖图
Our search space~(shown in Figure~\ref{fig_search_space}(b)) treats edge as entity-equivalent and adopts dual updating graph as architecture topology to learn the high-order representations of entity and edge, which presents more generic message passing mechanisms. 
Specifically, it models multi levels of entity and edge features, and allows complex dependence among them rather than classical sequential structure. 
Further, we design two kind of atomic operations: (1) entity updating operations for aggregating neighboring messages guided by relation features, (2) edge updating operations for modeling higher order relations guided by entity features. 
Based on the differentiable search algorithm, we can efficiently find the optimal dependence graph and updating operations to construct final GNN. 
%Our proposed edge-featured search space preserves different levels of entity and relation features, that are alternately updated to improve representation level via message passing mechanism. 
Experiments at six benchmarks show our EGNAS can search better GNNs with higher performance than current state-of-the-art methods, including hand-designed and searched-based GNNs. 

Our contributions can be summarized as follows:
\begin{itemize}
  \item We present edge-featured graph neural architecture search to find the optimal graph neural architecture, where a novel search space is designed with rich entity and edge updating operations. 
  %\item This is the first effort to explicitly introduce edge features and their complex dependence graph in search space to guide message passing. 
  \item This is the first effort that explicitly incorporates edge features into graph search space, where dual updating graph is introduced to explore complex entity/edge feature dependence and learn high-order representations, presenting more generic message passing mechanisms. 
  %\item This is the first effort that explicitly introducing dual updating graph in search space to jointly learn high-order entity and edge features, where complex feature dependence are explored for finding generic message passing mechanisms. 
  \item We evaluate EGNAS on six benchmarks at three typical graph tasks. The results show that graph architectures searched by EGNAS outperform the human-invented and search-based architectures. 
\end{itemize}

% 1. NAS + GNAS
% 2. CNN，RNN 通用的搜索空间，GNN无，特别之处。 GNN 的作用，核心是，层数设计 
% 3. 当前GNAS 介绍一下，总结GNAS如何对待节点和边。而近来edge-featured GNN 出现 意义所在（1，2句话）。
% 4. 把edge featured 用好， 这些GNNs工作怎么做 ， 难以融入到搜索机制中 ， 之前 GNAS 解决不了， 需要设计新的空间
% 5. 本文提出的方法。

% 1: NAS -> 图数据和GNN(边 关系建模) -> GNAS 只关注点而未关注边 -> 我们认为GNAS非常有潜力，自然解决这个问题 
% 我们认为通过NAS的方式建模和挖掘edge feature是很有潜力的。

% 2: 
\section{Related Work}

\textbf{Graph Neural Networks.}
GNNs have been successfully applied to operate on the graph-structure data~\cite{bresson2017residual, brockschmidt2020gnn, corso2020principal, kipf2016semi, li2015gated, NEURIPS2020_1534b76d, velivckovic2017graph, vignac2020building, xu2018powerful, yang2020factorizable}. 
Current GNNs are constructed on message passing mechanisms and can be categorized into two groups, isotropic and anisotropic. 
%! 选词 are based on 
Isotropic GNNs~\cite{hamilton2017inductive, kipf2016semi, li2015gated, vignac2020building, xu2018powerful} aim at extending the original convolution operation to graphs. 
Anisotropic GNNs enhance the original models with anisotropic operations on graphs~\cite{perona1990scale}, such as gating and attention mechanism~\cite{battaglia2016interaction, bresson2017residual, marcheggiani2017encoding, monti2017geometric, velivckovic2017graph, yang2020factorizable}. 
Anisotropic methods usually achieve better performance, since they can adaptively weight the edges according to the entity features and guide the message passing. 
However, these methods over-emphasize on entity rather than edge features. 
% 结论性的话

\textbf{Exploiting Edge Features.} 
Recently, researchers~\cite{chen2021edge, jiang2019censnet, Li_2019_CVPR, schlichtkrull2018modeling, yang2020nenn} have tried to intergrate edge features into GNN architecture. 
Schlichtkrull et al.~\cite{schlichtkrull2018modeling} propose R-GCNs to model relational data by grouping edges on graph, which indicates the edges cannot include continuous attributes. 
Works~\cite{chen2021edge, jiang2019censnet, yang2020nenn} assign high-dimensional feature for edges and iteratively update edge features using the same way as entity updating. 
These methods take no account of the dependencies between multi-level relation features and lack of diversified update functions to compute the edge features. 
% 不充分地利用了边地信息

\textbf{Neural Architecture Search.}
NAS aims at finding the optimal nerural architectures specific to dataset~\cite{baker2016designing, cai2018proxylessnas,chu2020darts,li2020sgas,liang2019darts+,liu2018darts,pham2018efficient,xie2018snas,zoph2016neural}. 
Search space and search strategy are the most essential components in NAS. 
Search space defines which architectures can be represented in principal. 
Search strategy details how to explore the search space. 
Methods can be mainly categorized into three groups, reinforcement learning (RL)~\cite{baker2016designing, zoph2016neural, zoph2018learning}, evolutionary algorithms (EA)~\cite{liu2017hierarchical, real2019regularized, real2017large} and gradient-based (GB)~\cite{liu2018darts, xu2019pc, zela2019understanding}. 
Due to the success in CNNs and RNNs, recent researchers~\cite{cai2021rethinking, gao2020graph, jiang2020graph, lai2020policy, nunes2020neural, zhang2021automated, zhou2019auto} apply NAS in graph machine learning domain to automatically design GNNs, termed as graph neural architecture search. 
Most of these works mainly focus on designing graph search space that can process non-euclidean data. 
However, during designing search space, they all ignore the information associated with edges that can be beneficial to graph tasks. 
To our best knowledge, we are the first to explicitly model edge features in graph NAS problem. 

% GNAS 结论性的话 对于edge-featured 是空白

\begin{figure*}[t]
    \centering
    \includegraphics[scale=0.42, trim = 0 0 0 0, clip]{ 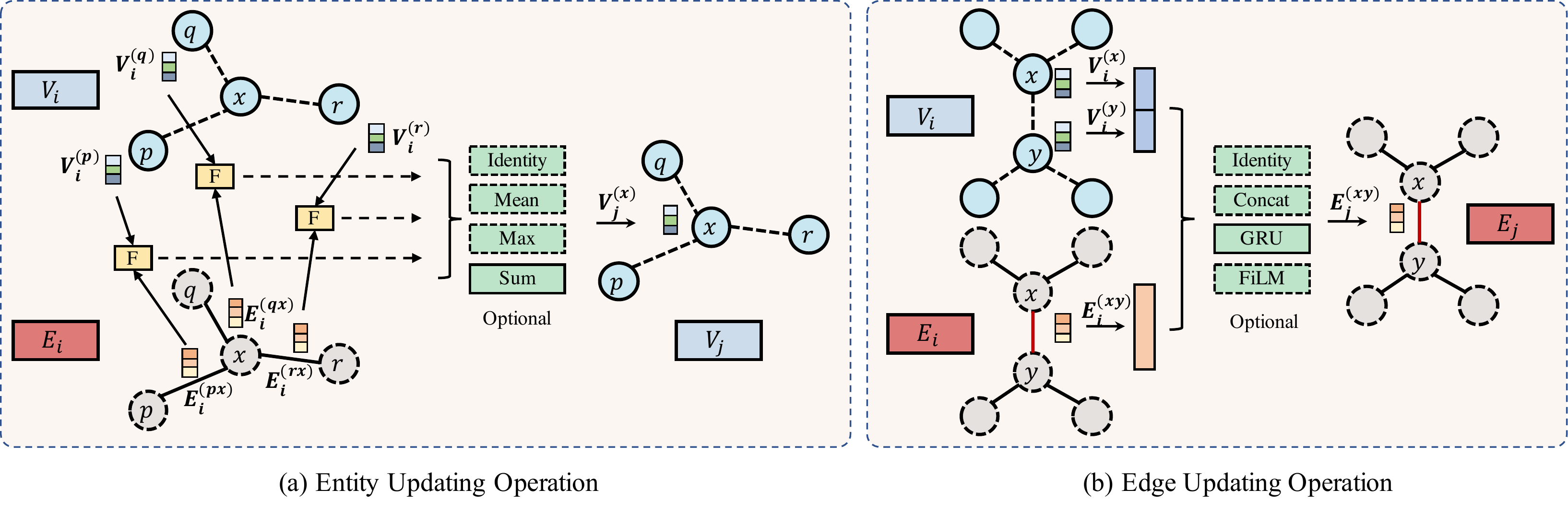}
    \caption{
        Illustration of entity updating operation and edge updating operation. 
        ``F'' module in subfigure (a) is used for processing messages based on feature-wise linear modulating operation before aggregating. 
    }
    \label{fig_update_operation}
\end{figure*}
\section{Method}
In this section, we first formulate the problem of graph neural architecture search, then introduce edge-featured graph search space and entity/edge updating operations. 
Next, we discuss the gradient-based search algorithm. 
Finally, we describe how to design final layers based on the specific task. 
\subsection{Problem Statement}
We formally define the graph neural architecture search as following bi-level optimization problem. 
Given graph search space $\mathcal{A}$, we aim to find the optimal GNN architecture $\alpha^{*} \in \mathcal{A}$ that minimizes the validation loss $\mathcal{L}_{val}(w^{*}(\alpha), \alpha^{*})$. 
The trainable weights $w^{*}$ associated with the architecture are obtained by minimizing the training loss $\mathcal{L}_{train}(w^{*}, \alpha^{*})$. 
Mathematically, it is written as follows. 
\begin{align}
    {\alpha}^{*} &= \mathop{argmin}\limits_{\alpha \in \mathcal{A}} \mathcal{L}_{val}( w^{*}(\alpha), \alpha ) \\
    w^{*}(\alpha) &= \mathop{argmin}\limits_{w} \mathcal{L}_{train}(w, \alpha)
\end{align}
The characteristic of GNN search problem is that the search space $\mathcal{A}$ consists of atomic operations which are designed to process complex graph-structure data and convey rich message passing mechanisms. 
\subsection{Search Space}
% explicitly models deep relational information as edge features 
Similar to the search space of CNNs, we search computational cells as the building blocks and stack them for the final model (shown in Figure~\ref{fig_search_space}(b)). In our edge-featured search space, each computational cell consists of dual directed acyclic graphs, termed as entity-updating graph and edge-updating graph. In the following, we will describe how the two graphs are constructed.

\textbf{Entity-updating Graph.} 
Entity-updating graph consists of an ordered sequence of $N$ nodes. Each node $\bm{V}_{i}$ is latent representations (\emph{i.e.} entity embeddings in a GNN layer) and each directed edge $(i,j)$ is associated with some entity-updating operation $o_{\mathcal{V}}^{(i,j)}$ that transforms $\bm{V}_{i}$ to $\bm{V}_{j}$ guided by relation features $\bm{E}_{i}$ (shown in Figure\ref{fig_update_operation}(a)). We compute each imtermediate node based on all its predecessors, written as
\begin{equation}
    \bm{V}_{j} = \sum_{0 \le i < j} o_{\mathcal{V} }^{(i,j)} (  \bm{V}_{i}, \bm{E}_{i}),
\end{equation}
where $1 \leq j \leq N$, $\bm{V}_{0}$ denotes the input entity representation. 
%, $\mathcal{F}(\cdot, \cdot)$ is feature-wise linear modulation. 
Inspired by works~\cite{brockschmidt2020gnn, perez2018film}, we use relation representation as input that determines an element-wise affine transformation of incoming messages. 
This allows the model to dynamically up-weight and down-weight features based on the information of relation. 
It yields the following update rule, 
\begin{gather}
    \bm{\beta}^{(s,t)}, \bm{\gamma}^{(s,t)} = g(\bm{E}^{(s,t)}_{i} ; \bm{\theta}_{i}), \\
    \bm{V}_{j}^{(t)} = \mathcal{M}(\{\bm{\gamma}^{(s,t)} \odot \bm{V}_{j}^{(s)} + \bm{\beta}^{(s,t)} | s \in \mathcal{N}(t)\}),
\end{gather}
where $g(\cdot)$ is a function to compute affine transformation, $\bm{\theta}_{i}$ is learnable parameters, $\mathcal{N}(t)$ denotes neighboring entities of target entity $t$. 
We define optional message aggregator $\mathcal{M}(\cdot)$ as a continuous function of multisets that aggregates messages on neighboring entities, such as $\mathop{sum}$, $\mathop{mean}$ and $\mathop{max}$. 
Different aggregators capture different types of information, work~\cite{cai2021rethinking} demonstrates that $\mathop{sum}$, $\mathop{mean}$, and $\mathop{max}$ do well in capturing structural, statistical and representative information from neighboring entities, respectively. 
Our search space allows to select the most appropriate message aggregators for different graph datasets. 
Besides, we introduce \emph{entity skip-connect} operation in search space to alleviate the gradient vanishing and over-smoothing problems. 
The output of the whole entity-updating graph is obtained by applying a reduction operation (\emph{e.g.} concatenation) to all the intermediate nodes. 

\textbf{Edge-updating Graph.} Edge-updating graph is a directed acyclic graph with the same number of nodes but different topology as entity-updating graph. 
Each node $\bm{E}_{i}$ is the latent relation representation and each directed edge $(i, j)$ is associated with some edge-updating operation $o^{(i,j)}_{\mathcal{E}}$ that learns higher order relation representation $\bm{E}_{j}$ from both $\bm{E}_{i}$ and its corresponding entity representation $\bm{V}_{i}$ (shown in Figure\ref{fig_update_operation}(b)). 
Similarly, the intermediate node in edge-updating graph follows below update rule,
\begin{equation}
\bm{E}_{j} = \sum_{0 \le i < j} o_{\mathcal{E} }^{(i,j)} ( \bm{E}_{i}, \bm{V}_{i} ),
\end{equation}
where $1 \leq j \leq N$, $\bm{E}_{0}$ denotes the input relation representation. 
Note that, if there is no original relation representation in datasets, then $\bm{E}_0$ is initialized with identity vectors $\bm{1} = [1]$. 
Specifically, for two entities $s$ and $t$, we first compute the joint entity representation as temporary relation feature $[\bm{V}^{(s)}_{i} \parallel \bm{V}^{(t)}_{i}]$, then use it to update current relation representation $\bm{E}^{(s,t)}_{i}$:
\begin{equation}
    \bm{E}^{(s,t)}_{j} = \mathcal{U}( \bm{E}^{(s,t)}_{i}, [ \bm{V}^{(s)}_{i} \parallel \bm{V}^{(t)}_{i} ])
\end{equation}
where $\parallel$ denotes concatenation operation, $\mathcal{U}(\cdot, \cdot)$ is an optional updating function. 
To present more message passing mechanisms, we introduce several update functions with different capabilities, such as \emph{Concat}, \emph{FiLM}~\cite{perez2018film} and \emph{GRU}~\cite{chung2014empirical}. 
\emph{Concat} denotes concatenation, that is the naivest operation of feature aggregation. 
\emph{FiLM} is short for feature-wise linear modulation that dynamically up-weight or down-weight old relation feature guided by temporary relation feature. 
\emph{GRU} adaptively determines how much old relation feature to forget and how much temporary relation features to inject. 
This is critical for preserving long-term relation and constructing robust high-order relation features. 
Besides, we also introduce a special operation \emph{edge skip-connect} to improve learning deep relation features. 
The output of the edge-updating graph is also obtained by applying a reduction operation to all the intermediate nodes. 
\footnote{All the candidate operations and computation details can be found in supplementary material.}

\subsection{Search Strategy}
Following previous works~\cite{cai2021rethinking, li2020sgas}, we adopt differentiable NAS strategy to find the optimal GNN. Instead of optimizing different operations separately, differentiable strategy maintains a supernet (one-shot model) containing all candidate operations, formally written as 
\begin{equation}
    \bar{o}^{(i,j)}(x) = \sum_{o \in \mathcal{O}} \frac{exp(\alpha^{(i,j)}_{o})}{\sum_{o' \in \mathcal{O}}exp(\alpha^{(i,j)}_{o'})} o(x),
\end{equation}
where $\mathcal{O}$ is a set of candidate operations, $\alpha^{(i,j)}$ is learnable vector that controls which operation is selected. Briefly, each mixed operation $\bar{o}^{(i,j)}$ is regarded as a probability distribution of all possible operations.
In this way, we can use gradient-based algorithms to jointly optimize the architecture and model weights. 
At the end of search, a discrete architecture is derived from supernet by replacing each mixed operation $\bar{o}^{(i,j)}$ with the most likely operation, i.e., $o^{(i,j)} = \mathop{argmax}_{o \in \mathcal{O}} \alpha^{(i,j)}_o$. 
For each node in cell, we retain at most two edges from all of its incoming edges. 
$\mathcal{O}$ can either be entity updating operation set $\mathcal{O}_\mathcal{V}$ or edge updating operation set $\mathcal{O}_\mathcal{E}$.

\subsection{Task-based Layer}
We design the final network layers depending on the specific task. 
Let $\mathcal{V} = \{v_1, v_2, ..., v_{|\mathcal{V}|}\}$ be the set of entities and $\mathcal{E} \subseteq \mathcal{V} \times \mathcal{V}$ be the set of edges, $\bm{V}_{o} \in \mathbb{R}^{|\mathcal{V}| \times d_\mathcal{V}}, \bm{E}_{o} \in \mathbb{R}^{|\mathcal{E}| \times d_\mathcal{E}}$ be the final entity representation and edge (relation) representation, respectively. 

\textbf{Node-level task layer}. 
For node classification task, the prediction $\bm{y}^{(i)}_{\mathcal{V}}$ is done as follows:
\begin{equation}
    \bm{y}^{(i)}_{\mathcal{V}} = \bm{P}_{\mathcal{V}}\bm{V}^{(i)}_{o}
\end{equation} 
where $\bm{P}_{\mathcal{V}} \in \mathbb{R}^{C \times d_{\mathcal{V}}}$ is the classifier, $C$ is the number of entity classes.

\textbf{Edge-level task layer}. 
For edge classification task, our method naturally makes predictions based on deep edge features $\bm{E}_{o}$, formally written as
\begin{equation}
    \bm{y}^{(s,t)}_{\mathcal{E}} = \bm{P}_{\mathcal{E}}\bm{E}^{(s,t)}_{o}
\end{equation}
where $\bm{P}_{\mathcal{E}} \in \mathbb{R}^{C \times d_{\mathcal{E}}}$ is a learnable matrix, $C$ is the number of edge classes. 
This is better than traditional GNN works~\cite{kipf2016semi, xu2018powerful, velivckovic2017graph} that concatenate the entity features as edge features, since the independent edge features are more discriminative at edge-level tasks. 

\textbf{Graph-level task layer}. 
For graph classification and regression tasks, we first use mean-pooling readout operation to compute global entity feature and global edge feature, then concatenates them as global graph representation. 
The prediction $\bm{y}_{\mathcal{G}}$ is computed as follows
\begin{equation}
    \bm{y}_{\mathcal{G}} = \bm{P}_{\mathcal{G}} \big[ \frac{1}{|\mathcal{V}|} \sum_{i \in \mathcal{V}} \bm{V}^{(i)}_{o} \big| \big| \frac{1}{|\mathcal{E}|} \sum_{(s,t) \in \mathcal{E}} \bm{E}^{(s,t)}_{o} \big],
\end{equation}
where $\bm{P}_{\mathcal{G}} \in \mathbb{R}^{C \times (d_{\mathcal{V}} + d_{\mathcal{E}})}$ is a learnable transformation matrix, $C$ is the number of graph classes. 

\section{Experiments}

\subsection{Experimental Setups}
\label{exp}
\textbf{Datasets.} 
\begin{table}[t]
    \small
    \centering
    \setlength{\tabcolsep}{3.2 mm}
    \renewcommand\arraystretch{0.8}
    \caption{Summary statistics of datasets. $\mathcal{V}, \mathcal{E}, \mathcal{G}$ denote node-level, edge-level and graph-level tasks.}
    \label{tab_summary}
    \begin{tabular}{l|cc|c|ccc}
    \toprule
                       & PATTERN     & CLUSTER     & TSP         & ZINC        & MNIST       & CIFAR10     \\ \midrule
    \#Graphs      & 14K         & 12K         & 12K         & 12K         & 70K         & 60K         \\
    \#Nodes       & 44-188      & 41-190      & 50-500      & 9-37        & 40-75       & 85-150      \\
    Total \#Nodes & 1,664,491   & 1,406,436   & 3,309,140   & 277,864     & 4,939,668   & 7,058,005   \\
    Task Level    &$\mathcal{V}$&$\mathcal{V}$&$\mathcal{E}$&$\mathcal{G}$&$\mathcal{G}$&$\mathcal{G}$\\ 
    \bottomrule
    \end{tabular}
    \label{tab_statistic}
\end{table}
We evaluate our method on six datasets (TSP, ZINC~\cite{irwin2012zinc}, MNIST~\cite{lecun1998gradient}, CIFAR10, PATTERN~\cite{dwivedi2020benchmarking} and CLUSTER~\cite{dwivedi2020benchmarking})
across three different level tasks (edge-level, graph-level and node-level). 
TSP dataset is based on the classical \emph{Travelling Salesman Problem}, which tests edge classification on 2D Euclidean graphs to indentify edges belonging to the optimal TSP solution. 
ZINC is one popular real-world molecular dataset of 250K graphs, whose task is graph property regression, out of which we select 12K for efficiency following works~\cite{cai2021rethinking, corso2020principal, dwivedi2020benchmarking}. 
MNIST and CIFAR10 are original classical image classification datasets and converted into graphs using superpixel~\cite{achanta2012slic} algorithm to test graph classification task. 
PATTERN and CLUSTER are node classification tasks generated via Stochastic Block Models~\cite{abbe2017community}, which are used to model communications in social networks by modulating the intra-community and extra-community connections. 
Details about the six datasets are shown in Table~\ref{tab_summary}, where $\mathcal{V}, \mathcal{E}, \mathcal{G}$ denote node-level, edge-level and graph-level tasks, respectively.

\textbf{Searching settings.}
In EGNAS, we define entity updating operation set $\mathcal{O}_{\mathcal{V}}$: \emph{sum}, \emph{max}, \emph{mean}, \emph{entity skip-connect} and \emph{zero}, edge updating operation set $\mathcal{O}_{\mathcal{E}}$: \emph{Concat}, \emph{GRU}, \emph{FiLM}, \emph{edge skip-connect} and \emph{zero}. 
Unless otherwise stated, each operation except \emph{skip-connect} and \emph{zero} is followed by a FC-ReLU-BN module. 
For fair comparison, on MNIST and CIFAR10, the search space consists of one computational cell. 
For other datasets, the search space consists of 4 cells. 
Each cell consists of 4 entity features and 4 edge features. 
In order to stablelize the gradient, additional residual connections are introduced in each computational cell. 
To carry out the architecture search, we hold out half of the training data as validation set. 
The supernet is trained using EGNAS for $40$ epochs, with batch size $64$ (for both the training and validation set). 
We use momentum SGD to optimize the weights $\bm{w}$, with initial learning rate $\eta_{\bm{w}}=0.025$ (anneald down to zero following a cosine schedule without restart), momentum $0.9$, and weight decay $3 \times 10^{-4}$. 
We use Adam~\cite{kingma2014adam} as the optimizer for $\bm{\bm{\alpha}}$, with initial learning rate $\eta_{\alpha} = 3 \times 10^{-4}$, momentum $\beta = (0.5, 0.999)$ and weight decay $10^{-3}$. 

\textbf{Training settings.}
\label{gpu}
For fair comparison, we follow all training settings~(data splits, optimizer, metrics, etc) in work~\cite{cai2021rethinking, dwivedi2020benchmarking}. 
Specifically, we adopt Adam~\cite{kingma2014adam} with the same learning rate decay for all runs. 
The learning rate is initialized with $10^{-3}$, which is reduced by half if the validation loss stop decreasing after 10 epochs. 
During training on CIFAR10 and MNIST dataset, we set dropout to 0.2 to alleviate the overfitting. 
% According to the searching settings, the searched GNN architecture contains 16 entity updating operations with perceptive fields size of 16. 
For MNIST and CIFAR10 datasets, we report the results of state-of-the-art GNNs with 4 layers. 
According to the searching settings, for other datasets, the GNNs with 16 layers are compared. 
% For fair comparison, we only report the experimental results of state-of-the-art GNNs with 16 layers. 
These experiments~(including searching and training) are run on NVIDIA GTX 2080Ti. 
Additional details can be found in the supplementary materials.

\subsection{Results of Graph-level Task}
\begin{table}[h]
    \small
    \centering
    \setlength{\tabcolsep}{2.1 mm}
    \renewcommand\arraystretch{0.8}
    \caption{
        Comparision with the state-of-the-art methods on MNIST and CIFAR10 datasets at graph classification task. 
        We report the number of parameters, test accuracy and search cost for all methods. 
    }
    \label{tab_graph}
    \begin{tabular}{@{}l|ccc|ccc@{}}
    \toprule
                                            & \multicolumn{3}{c|}{\bf{MNIST}}                         & \multicolumn{3}{c}{\bf{CIFAR10}}           \\
    \bf{Architecture}                       & \bf{Params} & \bf{Test Acc.}        & \bf{Search Cost}  & \bf{Params} & \bf{Test Acc.} & \bf{Search Cost} \\
                                            & (\bf{M})    & (\%)                  & \bf{(GPU days)}   & (\bf{M})    & (\%)           & \bf{(GPU days)}   \\
    \midrule
    GCN~\cite{kipf2016semi}                 & $0.10$      & $90.71\pm0.22$        & manual            & $0.10$        & $56.34\pm0.38$          & manual \\
    GIN~\cite{xu2018powerful}               & $0.10$      & $96.49\pm0.25$        & manual            & $0.10$        & $55.26\pm1.53$          & manual \\
    GraphSage~\cite{hamilton2017inductive}  & $0.10$      & $97.31\pm0.10$        & manual            & $0.10$        & $65.77\pm0.31$          & manual \\
    GAT~\cite{velivckovic2017graph}         & $0.11$      & $95.54\pm0.21$        & manual            & $0.11$        & $64.22\pm0.46$          & manual \\
    GatedGCN~\cite{bresson2017residual}     & $0.10$      & $97.34\pm0.14$        & manual            & $0.10$        & $67.31\pm0.31$          & manual \\
    PNA(E)~\cite{corso2020principal}        & -           & $97.94\pm0.12$        & manual            & -             & $70.47\pm0.72$          & manual \\
    GNAS-RL~\cite{gao2020graph}             & $0.48$      & $93.80\pm0.10$        & 5                 & $0.48$        & $58.33\pm0.63$          & 5      \\
    GNAS-MP~\cite{cai2021rethinking}        & $0.39$      & $98.01\pm0.10$        & 0.25              & $0.43$        & $70.10\pm0.44$          & 0.30   \\
    \midrule 
    EGNAS (w/o E)                           & $0.17$      & $97.71\pm0.07$        & 0.21              & $0.20$        & $69.54\pm0.30$          & 0.40   \\
    EGNAS (E)                               & $0.33$      & $\bm{98.32\pm0.08}$   & 0.21              & $0.37$        & $\bm{73.23\pm0.35}$     & 0.40   \\
    \bottomrule
    \end{tabular}
\end{table}
To test the capacity of our method at graph-level tasks, we assess it on MNIST, CIFAR10~(both are graph classification task) and ZINC~(graph regression task) datasets. 
The experimental results are reported in Table~\ref{tab_graph} and Table~\ref{tab_ZINC_TSP}. 
We observe that, first, our method surpasses all the state-of-the-art hand-designed and search-based GNNs by a large margin~($17.5\%\uparrow$ on ZINC, $3.8\%\uparrow$ on CIFAR10, $0.3\%\uparrow$ on MNIST), which demonstrates the effectiveness of EGNAS. 
Compared with the previous GNAS approaches~\cite{cai2021rethinking,gao2020graph}, our EGNAS achieves a great performance improvement with fewer parameters. %, indicating that our designed search space presents more efficient message passing mechanisms. 
This indicates that our designed search space with more efficient atomic operations is better than traditional graph search space, and the searched GNN has more compact graph architecture. 
Second, the performance of EGNAS without edge features degrades significantly, proving that exploiting relation information in edges is critical for learning better graph representations. 
%Further, on CIFAR10 and MNIST datasets, which contain no initial edge features, 
Even on CIFAR10 and MNIST datasets whose edges only describe binary topological connections, our EGNAS still mines latent relation to guide better message passing on graph. 

\subsection{Results of Edge-level Task}
\begin{table}[h]
    \small
    \centering
    \setlength{\tabcolsep}{1.7 mm}
    \renewcommand\arraystretch{0.8}
    \caption{
        Comparision with the state-of-the-art GNNs on ZINC and TSP datasets. 
        We record the number of parameters and search cost for all runs. 
        Besides, we report MAE metric for ZINC dataset and F1 metric for TSP dataset. 
        Notably, lower MAE indicates better performance. 
    }
    \label{tab_ZINC_TSP}
    \begin{tabular}{@{}l|ccc|ccc@{}}
    \toprule
                                            & \multicolumn{3}{c|}{\bf{ZINC}}                          & \multicolumn{3}{c}{\bf{TSP}}                     \\
    \bf{Architecture}                       & \bf{Params} & \bf{Test MAE.}         & \bf{Search Cost}  & \bf{Params} & \bf{Test F1.}  & \bf{Search Cost}  \\
                                            & (\bf{M})    &  (\bf{mean} $\bm{\pm}$ \bf{std}) & \bf{(GPU days)}   & (\bf{M})    &(\bf{mean} $\bm{\pm}$ \bf{std})& \bf{(GPU days)}   \\
    \midrule
    GCN~\cite{kipf2016semi}                 & $0.50$      & $0.367\pm0.011$        & manual            & $0.10$        & $0.630\pm0.001$         & manual \\
    GIN~\cite{xu2018powerful}               & $0.51$      & $0.526\pm0.051$        & manual            & $0.10$        & $0.656\pm0.003$         & manual \\
    GraphSage~\cite{hamilton2017inductive}  & $0.51$      & $0.398\pm0.002$        & manual            & $0.10$        & $0.665\pm0.003$         & manual \\
    GAT(E)~\cite{velivckovic2017graph}      & $0.53$      & $0.384\pm0.007$        & manual            & $0.10$        & $0.782\pm0.006$         & manual \\
    GatedGCN(E)~\cite{bresson2017residual}  & $0.51$      & $0.214\pm0.013$        & manual            & $0.53$        & $0.838\pm0.002$         & manual \\
    PNA(E)~\cite{corso2020principal}        & -           & $0.188\pm0.004$        & manual            & -             & -                       & -      \\
    GNAS-RL~\cite{gao2020graph}             & $1.07$      & $0.413$                & 5                 & -             & -                       & -      \\
    GNAS-MP~\cite{cai2021rethinking}        & $1.20$      & $0.242$                & 0.05              & $1.20$        & $0.742\pm0.02$          & 0.50   \\
    \midrule 
    EGNAS (w/o E)                           & $0.36$      & $0.196\pm0.003$        & 0.08              & $0.35$        & $0.685\pm0.003$         & 0.42   \\
    EGNAS (E)                               & $0.69$      & $\bm{0.150\pm0.005}$   & 0.08              & $0.68$        & $\bm{0.849\pm0.001}$    & 0.42   \\
    \bottomrule
    \end{tabular}
\end{table}
We compare the performance of our optimal searched architecture with state-of-the-art hand-designed and search-based GNNs on TSP dataset. 
The results are presented in Table~\ref{tab_ZINC_TSP}. 
We observe that GNNs incorporating edge features significantly outperform those only focusing on learning entity features, where the latter~\cite{velivckovic2017graph, cai2021rethinking} obtain the representation of edges for final classification by concatenating entity features. 
This indicates that at the task of edge classification, assigning independent features to each edge can reduce its dependence on the entity features, and thus improve the discriminability of edge representations. 
In addition, the optimal GNN architecture discovered by our EGNAS surpasses the state-of-the-art GatedGCN~\cite{bresson2017residual}, which reflects the effectiveness of EGNAS at edge-level tasks. 
%The performance improvement benefits from that our designed search space introduces dual updating graph as architecture topology to present more generic message passing mechanisms. 
The performance improvement benefits from that our dual updating graph facilitates to explore more generic message passing mechanisms.

\subsection{Results of Node-level Task}
\begin{table}[h]
    \small
    \centering
    \setlength{\tabcolsep}{1.9 mm}
    \renewcommand\arraystretch{0.8}
    \caption{
        Comparision with the state-of-the-art GNNs on PATTERN and CLUSTER datasets at node classification task. 
        We report the number of parameters, test accuracy and search cost for all methods. 
    }
    \label{tab_node}
    \begin{tabular}{@{}l|ccc|ccc@{}}
    \toprule
                                            & \multicolumn{3}{c|}{\bf{PATTERN}}                       & \multicolumn{3}{c}{\bf{CLUSTER}}               \\
    \bf{Architecture}                       & \bf{Params} & \bf{Test Acc.}        & \bf{Search Cost}  & \bf{Params} & \bf{Test Acc.} & \bf{Search Cost}  \\
                                            & (\bf{M})    & (\%)                  & \bf{(GPU days)}   & (\bf{M})    & (\%)           & \bf{(GPU days)}   \\
    \midrule
    GCN~\cite{kipf2016semi}                 & $0.50$      & $71.89\pm0.33$        & manual            & $0.50$        & $68.50\pm0.98$          & manual \\
    GIN~\cite{xu2018powerful}               & $0.51$      & $85.38\pm0.14$        & manual            & $0.52$        & $64.72\pm1.55$          & manual \\
    GraphSage~\cite{hamilton2017inductive}  & $0.50$      & $50.49\pm0.01$        & manual            & $0.50$        & $63.84\pm0.11$          & manual \\
    GAT~\cite{velivckovic2017graph}         & $0.53$      & $78.27\pm0.18$        & manual            & $0.53$        & $70.59\pm0.45$          & manual \\
    GatedGCN~\cite{bresson2017residual}     & $0.50$      & $86.51\pm0.09$        & manual            & $0.50$        & $76.08\pm0.20$          & manual \\
    %GNAS-RL~\cite{gao2020graph}             & $0.48$      & $85.21\pm0.01$        & 5                 & $0.48$        & $52.61\pm0.22$          & 5      \\
    GNAS-MP~\cite{cai2021rethinking}        & $1.60$      & $\bm{86.85\pm0.10}$   & 0.10              & $1.61$        & $74.77\pm0.15$          & 0.10   \\
    \midrule 
    EGNAS (w/o E)                           & $0.24$      & $86.15\pm0.15$        & 0.14              & $0.31$        & $71.80\pm0.25$          & 0.12    \\
    EGNAS (E)                               & $0.44$      & $86.20\pm0.11$        & 0.14              & $0.61$        & $\bm{76.65\pm0.11}$     & 0.12    \\
    \bottomrule
    \end{tabular}
    \vspace{-0.5 em}
\end{table}
% To evaluate our method at node classification task, we compare it with state-of-the-art GNNs on PATTERN and CLUSTER datasets. 
% The comparison results are reported in Table~\ref{tab_node}. 
We show the results at node classification task on PATTERN and CLUSTER datasets in Table~\ref{tab_node}. 
PATTERN dataset tests the fundamental graph task of recognizing specific predetermined subgraphs, and the CLUSTER dataset aims at identifying community clusters, where structural information matters. 
Notably, the graphs in these datasets represent community networks, in which the edge only plays role in connecting two nodes. %, concealing few semantic information. 
% we suspect the message propagation through intra-community and extra-community connections should be different on CLUSTER. 
Interestingly, we find that our EGNAS still achieves competitive performance on PATTERN and surpasses all the state-of-the-art GNNs on CLUSTER. 
% The edges are heterogeneous in CLUSTER, where the message propagation through intra-community and extra-community connections should be different. 
Actually, the message propagation through intra-community and extra-community connections should be different on CLUSTER. 
% The GNNs searched by EGNAS have the potential to mine local structural similarities between nodes and can distinguish intra-community and extra-community connections. 
The GNNs searched by EGNAS can still distinguish intra-community and extra-community connections through mining local structural similarities between nodes. 
Specifically, they model structural similarity information in edge features using edge updating operations, to guide message passing and help identify specific communities on graph.

\subsection{Ablation Study for Architectures}
\begin{figure*}[h]
  \centering
  \includegraphics[scale=0.50, trim = 0 0 0 0, clip]{ 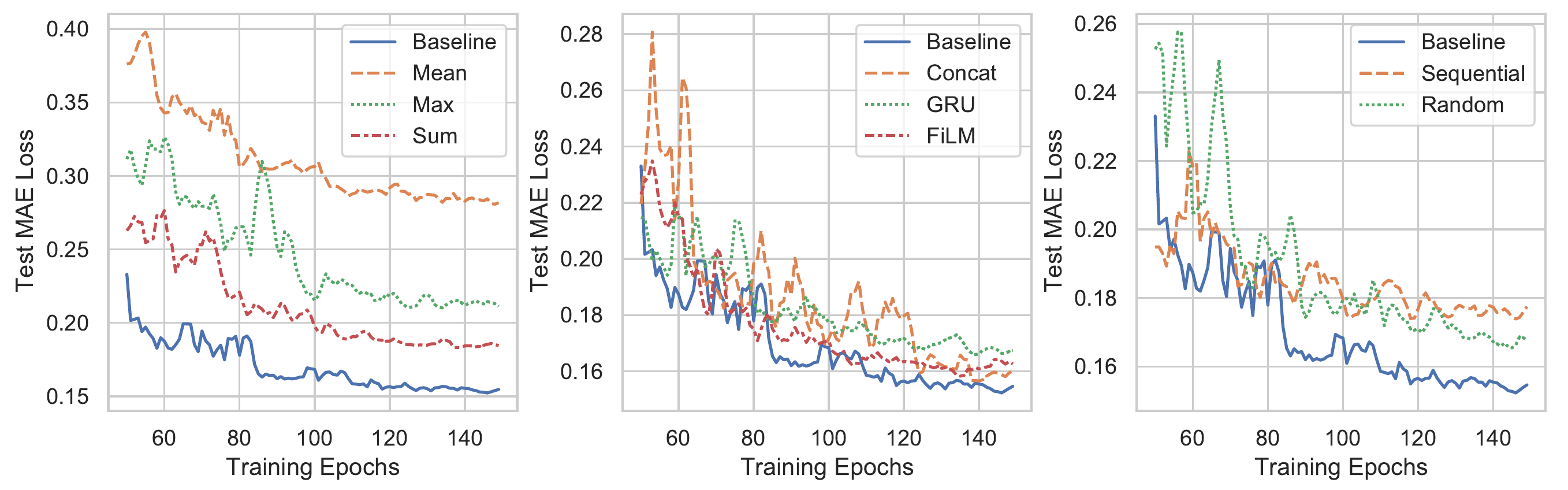}
  \caption{
    Experiments of modifying searched GNN architecture on ZINC dataset. 
    Baseline architecture is the best GNN architecture searched by the EGNAS, the others are obtained by modifying baseline architecture based on different rules, such as replacing entity updating operation~(left), edge updating operation~(middle) and modifying architecture topology~(right). 
    %! (, , ,)
    %Left figure shows the evaluation result where we turn all entity updating operations of the architecture into (Mean, Max, Sum), keeping other operations unchanged. 
  }
  \label{fig_modify}
\end{figure*}
Here, we explore what happens if we make some changes to the optimal searched architecture, which is termed as ``Baseline'' architecture.
%All the experiments are conducted on ZINC, a representative real-world dataset. 
First, we study the impact of the replacement of atomic operations. 
Specifically, we replace all the entity updating operations in ``Baseline'' architecture with a specific operation (\emph{e.g.} \emph{Mean}, \emph{Max}, \emph{Sum}), with other operations unchanged, obtaining three modified architectures. 
Similarly, we replace all the edge updating operations in ``Baseline'' architecture with \emph{Conat}, \emph{GRU} and \emph{FiLM} operations, respectively, for ablations of edge updating operations. 
These six architectures are retrained on ZINC dataset and the test performance in terms of MAE is shown in Figure~\ref{fig_modify}~(left and middle). 
We can observe that both convergence speed and MAE of the modified architectures are worse than the ``Baseline'' architecture, which means that the representation ability of a single feature updating operation is limited. %! 
Further, our method can automatically search the optimal way to combine different entity/edge updating operations, discovering novel message passing mechanisms and enhancing the graph representation learning capability of GNN. 

Second, we study the impact of architecture topology~(feature dependence graph). 
We randomly sample a graph neural architecture from edge-featured graph search space, termed as ``Random''. 
Besides, we obtain a ``Sequential'' architecture by removing some edges from edge-updating graph of ``Baseline'' architecture to make sure that $\bm{E}_{i}$ just depends on $\bm{E}_{i-1}$. 
Experimental results~(the right of Figure~\ref{fig_modify}) show that ``Random'' is better than ``Sequential'' but worse than ``Baseline'' architecture. 
This emphasizes that searching complex dependence graph of multi-level edge features matters. 
It is also interesting to note that randomly sampled architecture is competitive, which reflects the importance of the graph search space design.

\begin{figure*}[h]
  \centering
  \includegraphics[scale=0.52, trim = 10 0 0 0, clip]{ 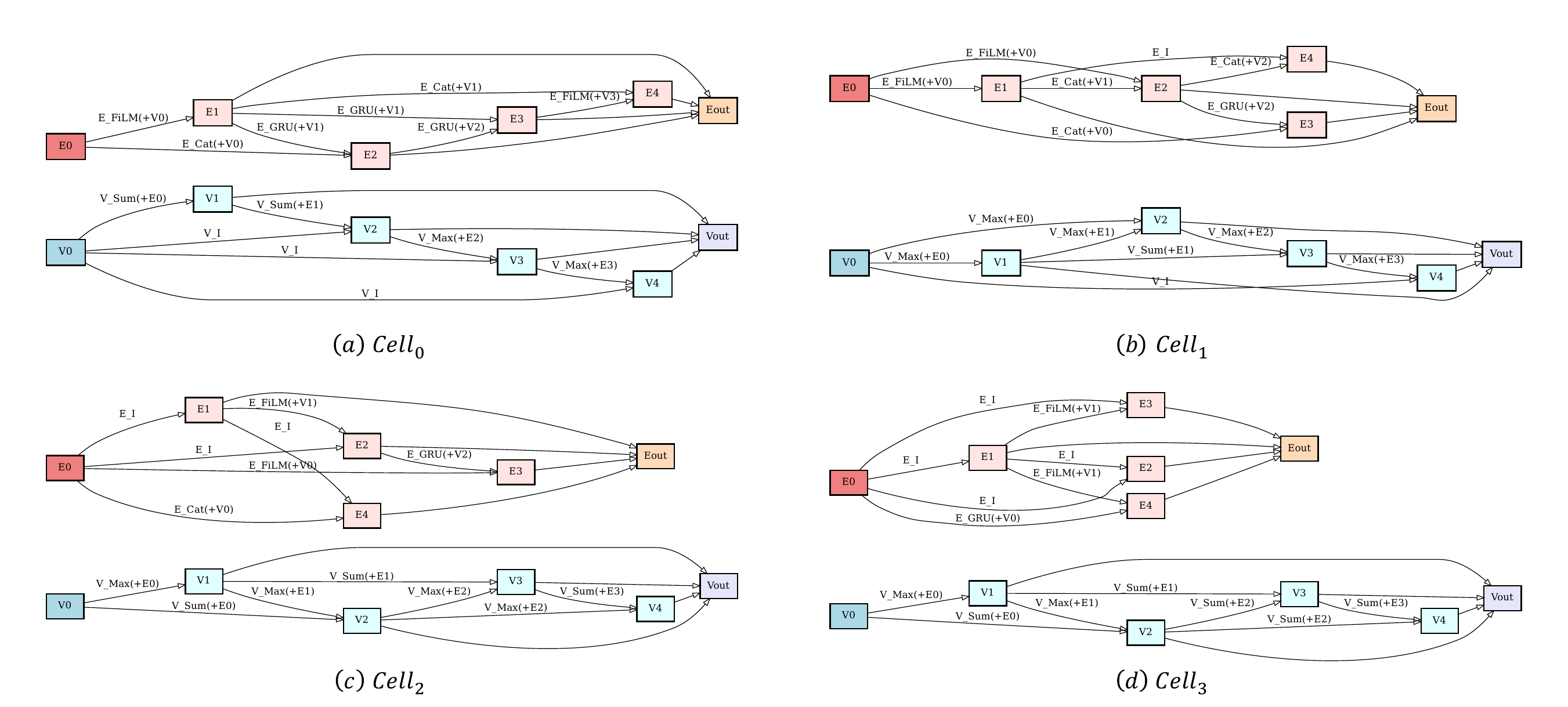}
  \caption{
    Best architecture with 4 cells searched on ZINC dataset. 
    Each subfigure corresponds to a specific computational cell, in which the above and below graphs describe edge-updating graph and entity-updating graph, respectively. 
    Notably, entity updating operations with $(+\bm{E}_{x})$ means the architecture updates entity features guided by edge features $\bm{E}_{x}$. 
    Similar, edge updating operations with $(+\bm{V}_{x})$ denotes that the entity features $\bm{V}_{x}$ are involved in the computation of learning edge features. 
  }
  \label{fig_archs}
\end{figure*}

\subsection{Further Discussion}
% According to the experiments, there are still some interesting phenomena about relation modeling that deserves further research. 
According to the experiments, we have observed some interesting phenomena that may inspire the design of edge-featured GNNs. 
% The topology of edge updating graph is quite different from that of entity updating graph. 
The longest path in the edge updating graph is getting shorter as the computational cell depth increases~(more nodes directly connect to $\bm{E}_{0}$ with \emph{edge skip-connect} operation). 
As depth increases, the edge features are smoothed along with entity features, weakening the ability to characterize the local structural similarity. 
Our EGNAS can automatically detect this phenomenon and supplement some lower order edge features via extra connections to counteract this smoothness. 
Another evidence is that the ``Sequential'' architecture performs worse than ``Baseline'' architecture~(shown in the right of Figure~\ref{fig_modify}), where ``Baseline'' contains more connections from lower order edge features in deeper computational cells. 
The searched architectures on other datasets also show similar preference. 
More results can be found in supplementary materials. 

\section{Conclusion}
In this paper, we incorporate edge features in graph neural architecture search for the first time, to find the optimal GNN architecture that can mine and exploit latent relation information concealed in edges. 
Specifically, we design edge-featured graph search space with novel atomic operations, which allows to explore complex entity/edge feature dependence and presents more generic message passing mechanisms. 
Experiments on six benchmarks at three classical graph tasks demonstrate that our EGNAS rivals all the human-invented and search-based methods. 
Further, we analyze the searched optimal architectures and confirm the effectiveness of our EGNAS. 
% Further, we analyze the searched optimal architectures and observes some preferences of architecture design. 
% We hope these findings may inspire further works on modeling relational information for graph-structure data analysis. 
\section*{Boarder Impact}
Our paper proposes edge-featured graph neural architecture search method that further introduces novel graph search space with more generic message passing mechanisms and find the optimal GNN architecture. 
Notably, this is a significant improvement on designing graph search space. 
It will have a direct impact on the application of graph neural networks in the industry. 
The positive impacts are described as follows. 
First, we can efficiently find better GNN architectures that can mine and exploit the relation information implied in graph. 
So the practical application of GNN to the corresponding area can be fostered. 
Second, we analyze the optimal searched GNN architectures and summarize some findings which may have implications for the design of GNN. 
Moreover, by applying our EGNAS to different tasks and datasets, we think that it could bring researchers expertise in understanding graph neural architectures. 
There are also some limitations in our method. 
For example, we mainly focus on designing generic graph search space rather than search strategy. 
We will develop specific search strategies for graph neural architecture search to improve the efficiency and performance of search in future works. 

\clearpage
% {
  \small
  \bibliographystyle{plain}
  % \bibliography{egbib}

% }

% \input{checklist}

\clearpage
\appendix
\section{Implementation Details}
In this section, we details how entity updating operations and edge updating operations are formulated. 
We declare that $\bm{V} \in \mathbb{R}^{|\mathcal{V}| \times d_{\mathcal{V}}}$ denotes entity feature, $\bm{E} \in \mathbb{R}^{|\mathcal{E}| \times d_{\mathcal{E}}}$ denotes edge feature, $\mathcal{V}$ is the set of entities, $\mathcal{E}$ is the set of edges, $d_{\mathcal{V}}$ is the dimension of entity feature, $d_{\mathcal{E}}$ is the dimension of edge feature. 

\subsection{Entity Updating Operation}
As discussed in body content, we use relation representation as input that determines an element-wise affine transformation of incoming messages, allowing the model to dynamically up-weight and down-weight features based on the information of relation. 
It yields the following update rule, 
\begin{gather}
    \bm{\beta}^{(s,t)}, \bm{\gamma}^{(s,t)} = g(\bm{E}^{(s,t)}_{i} ; \bm{\theta}_{i}), \\
    \bm{V}_{j}^{(t)} = \mathcal{M}(\{\bm{\gamma}^{(s,t)} \odot \bm{V}_{j}^{(s)} + \bm{\beta}^{(s,t)} | s \in \mathcal{N}(t)\}),
\end{gather}
where $g(\cdot)$ is a function to compute affine transformation, $\bm{\theta}_{i}$ is learnable parameters, $\mathcal{N}(t)$ denotes neighboring entities of target entity $t$, $\mathcal{M}$ is an optional neighbor aggregator. 
We define \emph{Max}, \emph{Mean}, \emph{Sum} entity updating operations for aggregating messages, with neighbor aggregator $\mathcal{M}(\cdot)$ as $max(\cdot)$, $mean(\cdot)$ and $sum(\cdot)$, respectively. 
Besides, we also introduce a special operation \emph{entity skip-connect} to alleviate the gradient vanishing and over-smoothing problems, formulated as 
\begin{equation}
  o_{\mathcal{V}\cdot skip}(\bm{V}_{i}, \bm{E}_{i}) = \bm{V}_{i}. 
\end{equation}

\subsection{Edge Updating Operation}
As we discussed in body content, the edge updating equation can be written as follows:
\begin{equation}
  \bm{E}^{(s,t)}_{j} = \mathcal{U}( \bm{E}^{(s,t)}_{i}, [ \bm{V}^{(s)}_{i} \parallel \bm{V}^{(t)}_{i} ])
\end{equation}
where $\parallel$ denotes the concatenation operation, $\mathcal{U}(\cdot, \cdot)$ is an optional updating function. 
Since we define multiple edge updating operations \emph{Concat}, \emph{GRU}, \emph{FiLM} and \emph{edge skip-connect}, the difference between them is the choice of the updating function $\mathcal{U}(\cdot, \cdot)$. 

\textbf{For \emph{Concat} Operation. }
\emph{Concat} denotes the concatenation, that is the naivest operation of feature aggregation. 
We use the following updating rule:
\begin{equation}
  \mathop{\mathcal{U}}\limits_{Concat}(\bm{E}^{(s,t)}_{i}, [ \bm{V}^{(s)}_{i} \parallel \bm{V}^{(t)}_{i} ]) = MLP([\bm{E}^{(s,t)}_{i} \parallel \bm{V}^{(s)}_{i} \parallel \bm{V}^{(t)}_{i}])
\end{equation}
where $MLP(\cdot)$ denotes a multi-layer perceptron that obtains new edge feature with $d_{\mathcal{E}}$ dimension. 

\textbf{For \emph{GRU} Operation. }
\emph{GRU}~\cite{chung2014empirical} adaptively determines how much old relation feature to forget and how much temporary relation features to inject. 
This is critical for preserving long-term relation and constructing robust high-order relation features. 
Mathematically, it can be formulated as
\begin{gather}
\bm{x} = ReLU(\bm{P}_{x}[ \bm{V}^{(s)}_{i} \parallel \bm{V}^{(t)}_{i} ]) \\
\bm{r} = \sigma(\bm{U}_{r}\bm{x} + \bm{W}_{r}\bm{E}^{(s,t)}_{i}) \\
\bm{z} = \sigma(\bm{U}_{z}\bm{x} + \bm{W}_{z}\bm{E}^{(s,t)}_{i}) \\
\bm{h} = tanh(\bm{U}_{h}\bm{x} + \bm{W}_{h}(\bm{r} \odot \bm{E}^{(s,t)}_{i})) \\
\mathop{\mathcal{U}}\limits_{GRU}(\bm{E}^{(s,t)}_{i}, [ \bm{V}^{(s)}_{i} \parallel \bm{V}^{(t)}_{i} ]) = (\bm{1} - \bm{z}) \odot \bm{E}^{(s,t)}_{i} + \bm{z} \odot \bm{h}
\end{gather}
where $\bm{P}_{x} \in \mathbb{R}^{d_{\mathcal{E}} \times 2d_{\mathcal{V}}}$, $\bm{U}_{r} \in \mathbb{R}^{d_{\mathcal{E}} \times d_{\mathcal{E}}}$, $\bm{W}_{r} \in \mathbb{R}^{d_{\mathcal{E}} \times d_{\mathcal{E}}}$, $\bm{U}_{z} \in \mathbb{R}^{d_{\mathcal{E}} \times d_{\mathcal{E}}}$, $\bm{W}_{z} \in \mathbb{R}^{d_{\mathcal{E}} \times d_{\mathcal{E}}}$, $\bm{U}_{h} \in \mathbb{R}^{d_{\mathcal{E}} \times d_{\mathcal{E}}}$ and $\bm{W}_{h} \in \mathbb{R}^{d_{\mathcal{E}} \times d_{\mathcal{E}}}$ are learnable matrices, $ReLU$ is ReLU activation function, $\sigma(\cdot)$ is sigmoid activation function. 

\textbf{For \emph{FiLM} Operation. }
\emph{FiLM} is short for feature-wise linear modulation~\cite{perez2018film} that dynamically up-weight or down-weight old relation feature guided by temporary relation feature. 
The required updating function can be written as
\begin{gather}
  \bm{\beta}^{(s,t)}, \bm{\gamma}^{(s,t)} = g([\bm{V}^{(s)}_{i} \parallel \bm{V}^{(t)}_{i}]; \bm{\theta}_{i}) \\
  \mathop{\mathcal{U}}\limits_{FiLM}(\bm{E}^{(s,t)}_{i}, [ \bm{V}^{(s)}_{i} \parallel \bm{V}^{(t)}_{i} ]) =
  \bm{\gamma}^{(s,t)} \odot \bm{E}^{(s,t)}_{i} + \bm{\beta}^{(s,t)}
\end{gather}
where $g(\cdot)$ is a function to compute affine transformation, $\bm{\theta}_{i}$ is learnable parameters. 

\textbf{For \emph{edge skip-connect} Operation. }
We introduce a special operation \emph{edge skip-connect} to improve learning deep relation features, formulated as 
\begin{gather}
  \mathop{\mathcal{U}}\limits_{skip-connect}(\bm{E}^{(s,t)}_{i}, [ \bm{V}^{(s)}_{i} \parallel \bm{V}^{(t)}_{i} ]) = \bm{E}^{(s,t)}_{i}
\end{gather}
The \emph{skip-connect} plays role in supplementing low level representations and alleviating gradient vanishing.

\section{Architecture Visualization}
\begin{figure*}[h]
  \centering
  \includegraphics[scale=0.51, trim = 10 0 0 0, clip]{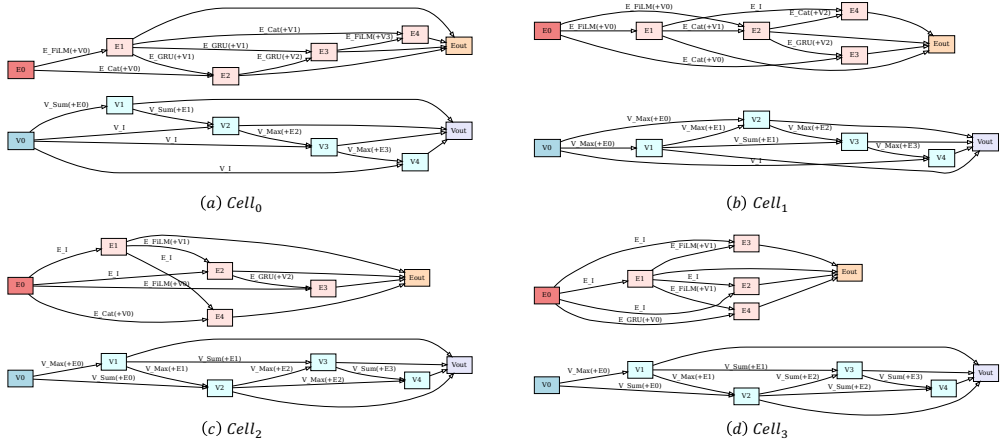}
  \caption{
    Best architecture with 4 cells searched on ZINC dataset. 
  }
  \label{fig_ZINC}
\end{figure*}
\begin{figure*}[t]
  \centering
  \includegraphics[scale=0.50, trim = 10 0 0 0, clip]{ 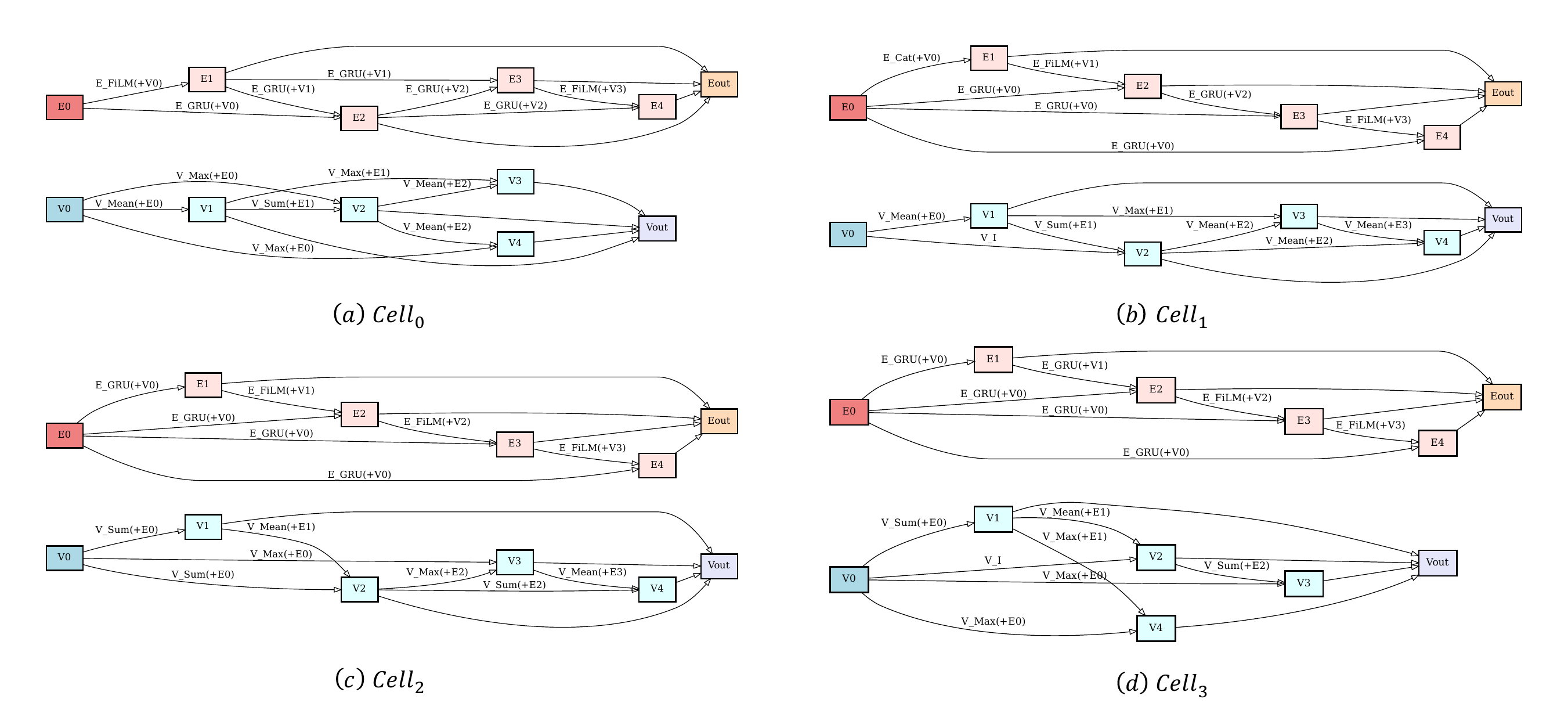}
  \caption{
    Best architecture with 4 cells searched on TSP dataset. 
  }
  \label{fig_TSP}
\end{figure*}
\begin{figure*}[t]
  \centering
  \includegraphics[scale=0.50, trim = 10 0 0 0, clip]{ 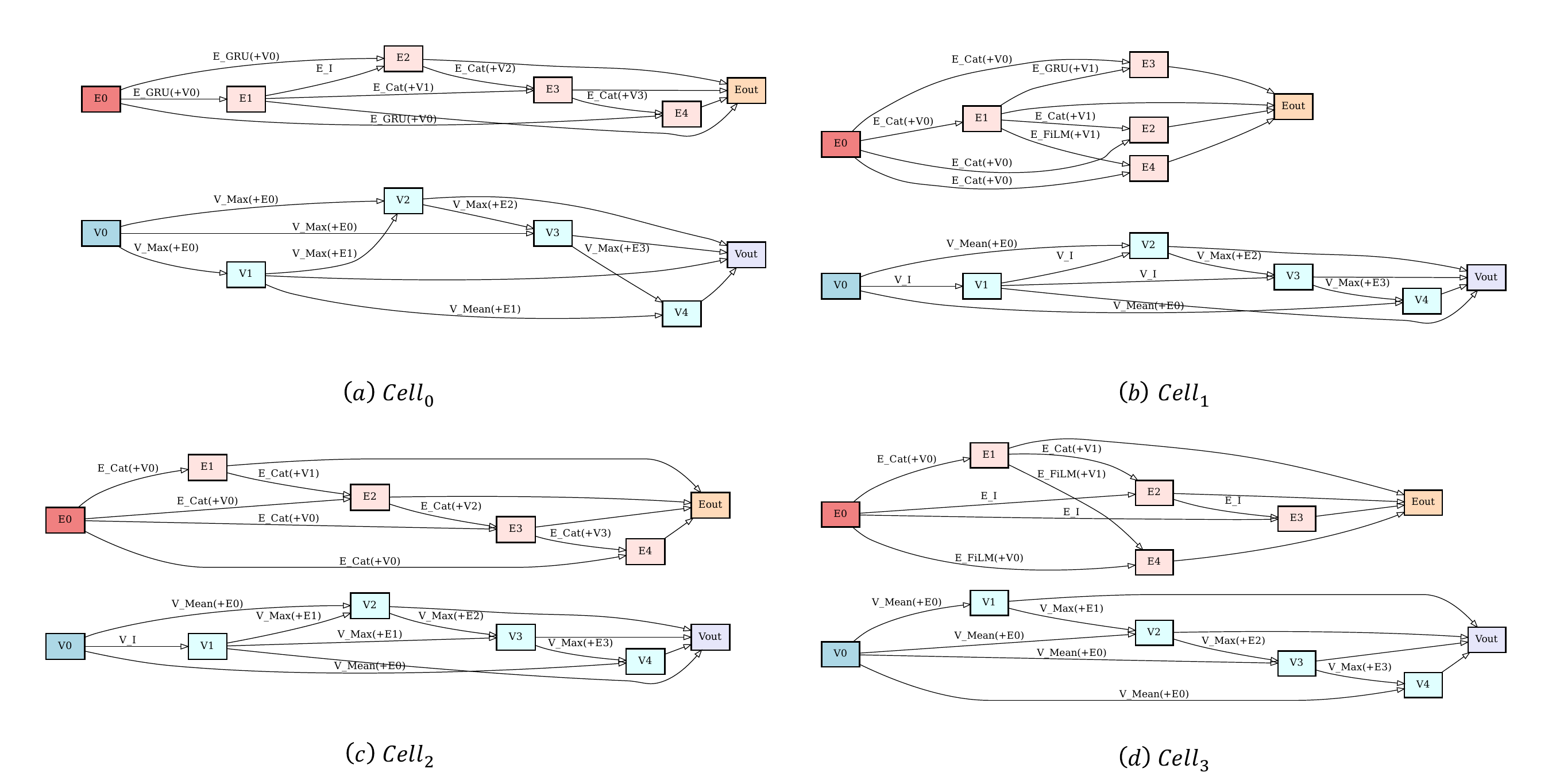}
  \caption{
    Best architecture with 4 cells searched on CLUSTER dataset. 
  }
  \label{fig_CLUSTER}
\end{figure*}
\begin{figure*}[t]
  \centering
  \includegraphics[scale=0.49, trim = 10 0 0 0, clip]{ 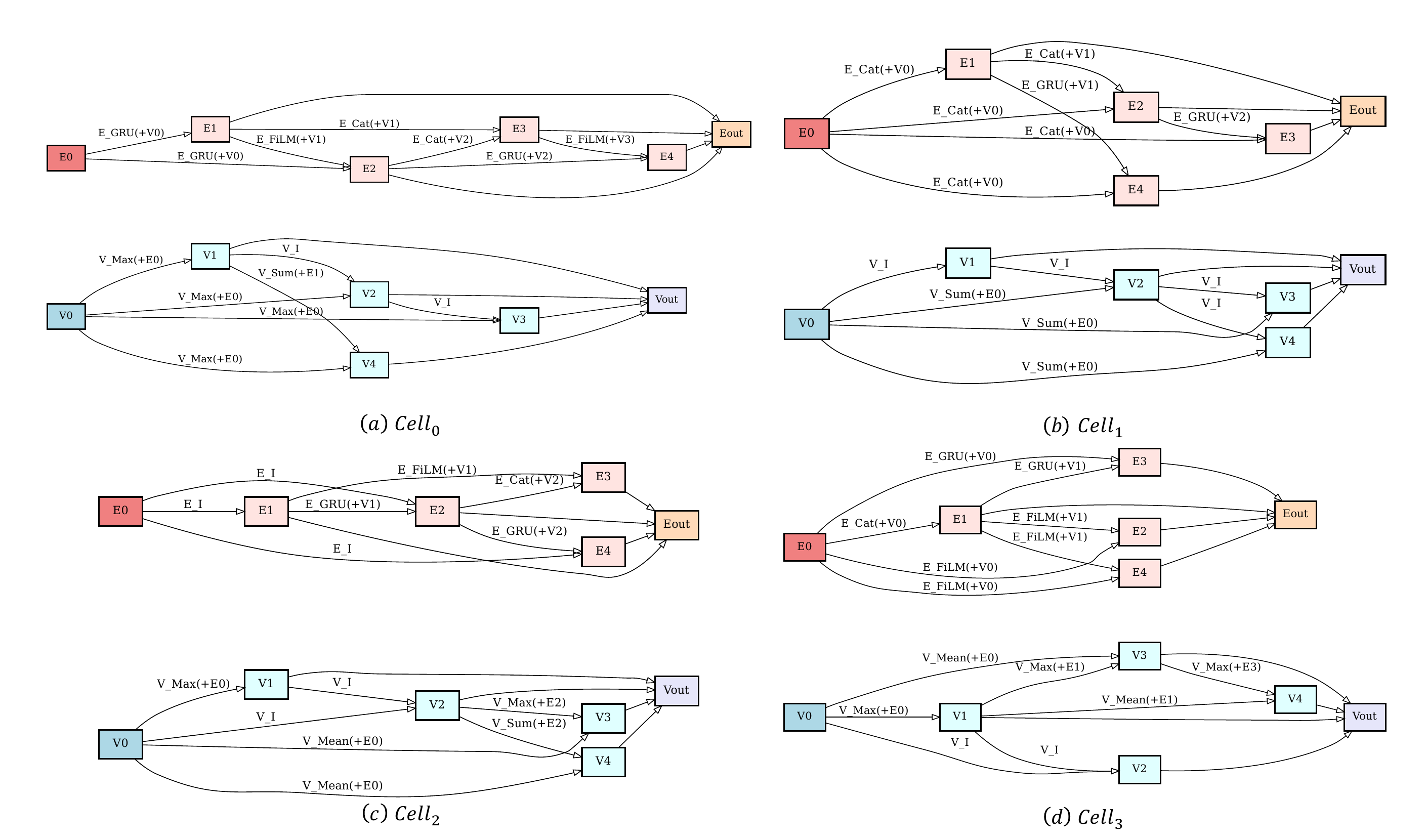}
  \caption{
    Best architecture with 4 cells searched on PATTERN dataset. 
  }
  \label{fig_PATTERN}
\end{figure*}
\begin{figure*}[t]
  \centering
  \includegraphics[scale=0.49, trim = 10 0 0 0, clip]{ 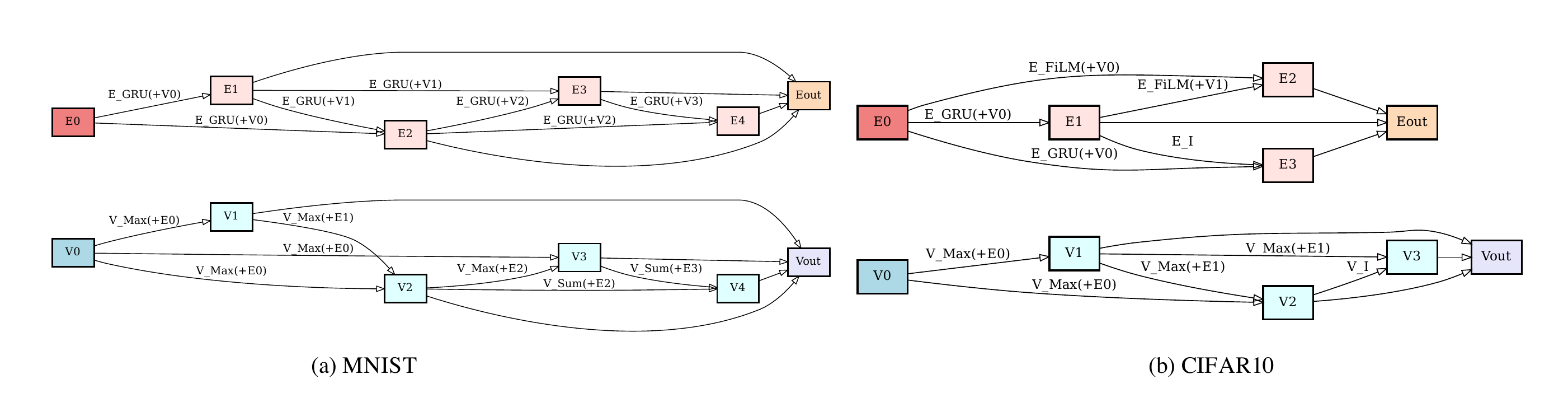}
  \caption{
    Best architecture with one cell searched on MNIST and CIFAR10 datasets. 
  }
  \label{fig_MNIST_CIFAR10}
\end{figure*}

As a supplementary for Section 4.6 in body content, we visualize the best searched architecture for each dataset, including ZINC~(shown in Figure~\ref{fig_ZINC}), TSP~(shown in Figure~\ref{fig_TSP}), CLUSTER~(shown in Figure~\ref{fig_CLUSTER}), PATTERN~(shown in Figure~\ref{fig_PATTERN}), MNIST~(shown in Figure~\ref{fig_MNIST_CIFAR10}[a]) and CIFAR10~(shown in Figure~\ref{fig_MNIST_CIFAR10}[b]). 
The visualization results on these datasets show similar preference as discussed in Section 4.6, \emph{i.e.}, as depth increases, the edge features are smoothed along with entity features, weakening the ability to characterize the local structural similarity. 
Our EGNAS can automatically detect this phenomenon and supplement some lower order edge features via extra connections to counteract this smoothness. 

\section{Extensive Experiments}
To evaluate the sensitivity of our method to the size of the receptive field, we conduct extensive experiments at edge-level task (on TSP dataset) and graph-level task (on ZINC dataset), then report the results in Table~\ref{tab_ZINC_TSP_ex}. 
Besides, we also conduct extensive experiments at node-level task (no both PATTERN and CLUSTER datasets) and report the results in Table~\ref{tab_node_ex}. 
We can find that, with the increase of receptive field, the performance of our model becomes better and better. 
Our method can achieve competitive results in any receptive field. 
In addition, through the ablation studies, we can see that incorporating edge features into search space is critical for improving performance of searched GNN architectures. 

\begin{table}[h]
    \small
    \centering
    \setlength{\tabcolsep}{1.2 mm}
    \renewcommand\arraystretch{1.0}
    \caption{
        Comparision with the state-of-the-art GNNs on ZINC and TSP datasets. 
        We record the number of parameters and search cost for all runs. 
        Besides, we report MAE metric for ZINC dataset, F1 metric and the size of receptive field $L$ for TSP dataset. 
        Notably, lower MAE indicates better performance. 
    }
    \label{tab_ZINC_TSP_ex}
    \begin{tabular}{@{}l|c|ccc|ccc@{}}
    \toprule
                                          &        & \multicolumn{3}{c|}{\bf{ZINC}}                          & \multicolumn{3}{c}{\bf{TSP}}                     \\
    \bf{Architecture}                     & \bf{L} & \bf{Params} & \bf{Test MAE.}         & \bf{Search Cost}  & \bf{Params} & \bf{Test F1.}  & \bf{Search Cost}  \\
                                          &        & (\bf{M})    &  (\bf{mean} $\bm{\pm}$ \bf{std}) & \bf{(GPU days)}   & (\bf{M})    &(\bf{mean} $\bm{\pm}$ \bf{std})& \bf{(GPU days)}   \\
    \midrule
    GCN~\cite{kipf2016semi}               &   4    & $0.10$      & $0.459\pm0.006$        & manual            & $0.10$        & $0.630\pm0.001$         & manual \\
    GCN~\cite{kipf2016semi}               &   16   & $0.50$      & $0.367\pm0.011$        & manual            & -             & -                       & - \\
    GIN~\cite{xu2018powerful}             &   4    & $0.10$      & $0.387\pm0.015$        & manual            & $0.10$        & $0.656\pm0.003$         & manual \\
    GIN~\cite{xu2018powerful}             &   16   & $0.51$      & $0.526\pm0.051$        & manual            & -             & -                       & - \\
    GraphSage~\cite{hamilton2017inductive}&   4    & $0.09$      & $0.468\pm0.003$        & manual            & $0.10$        & $0.665\pm0.003$         & manual \\
    GraphSage~\cite{hamilton2017inductive}&   16   & $0.51$      & $0.398\pm0.002$        & manual            & -             & -                       & - \\
    GAT~\cite{velivckovic2017graph}       &   4    & $0.10$      & $0.475\pm0.007$        & manual            & $0.10$        & $0.782\pm0.006$         & manual \\
    GAT~\cite{velivckovic2017graph}       &   16   & $0.53$      & $0.384\pm0.007$        & manual            & -             & -                       & - \\
    GatedGCN(E)~\cite{bresson2017residual}&   4    & $0.11$      & $0.375\pm0.003$        & manual            & $0.10$        & $0.808\pm0.003$         & manual \\
    GatedGCN(E)~\cite{bresson2017residual}&   16   & $0.51$      & $0.214\pm0.013$        & manual            & $0.53$        & $0.838\pm0.002$         & manual \\
    GNAS-RL~\cite{gao2020graph}           &   4    & $0.48$      & $0.480$                & 5                 & -             & -                       & -      \\
    GNAS-RL~\cite{gao2020graph}           &   16   & $1.07$      & $0.540$                & 5                 & -             & -                       & -      \\
    GNAS-MP~\cite{cai2021rethinking}      &   4    & $0.41$      & $0.276$                & 0.05              & $1.20$        & $0.742\pm0.02$          & 0.50   \\
    GNAS-MP~\cite{cai2021rethinking}      &   16   & $1.20$      & $0.260$                & 0.05              & -             & -                       & -      \\
    \midrule 
    EGNAS (w/o E)                         &   4    & $0.14$      & $0.448\pm0.005$        & 0.04              & $0.10$        & $0.645\pm0.005$         & 0.22   \\
    EGNAS (E)                             &   4    & $0.27$      & $0.228\pm0.010$        & 0.04              & $0.18$        & $0.816\pm0.002$         & 0.22   \\
    EGNAS (w/o E)                         &   16   & $0.36$      & $0.196\pm0.003$        & 0.08              & $0.35$        & $0.685\pm0.003$         & 0.42   \\
    EGNAS (E)                             &   16   & $0.69$      & $\bm{0.150\pm0.005}$   & 0.08              & $0.68$        & $\bm{0.849\pm0.001}$    & 0.42   \\
    \bottomrule
    \end{tabular}
\end{table}
\begin{table}[h]
    \small
    \centering
    \setlength{\tabcolsep}{1.2 mm}
    \renewcommand\arraystretch{1.0}
    \caption{
        Comparision with the state-of-the-art GNNs on PATTERN and CLUSTER datasets at node classification task. 
        We report the number of parameters, test accuracy, search cost and the size of receptive field $L$ for all methods. 
    }
    \label{tab_node_ex}
    \begin{tabular}{@{}l|c|ccc|ccc@{}}
    \toprule
                                           &      & \multicolumn{3}{c|}{\bf{PATTERN}}                       & \multicolumn{3}{c}{\bf{CLUSTER}}               \\
    \bf{Architecture}                      &\bf{L}& \bf{Params} & \bf{Test Acc.}        & \bf{Search Cost}  & \bf{Params} & \bf{Test Acc.} & \bf{Search Cost}  \\
                                           &      & (\bf{M})    & (\%)                  & \bf{(GPU days)}   & (\bf{M})    & (\%)           & \bf{(GPU days)}   \\
    \midrule
    GCN~\cite{kipf2016semi}                & 4    & $0.10$      & $50.52\pm0.00$        & manual            & $0.10$        & $20.94\pm0.00$          & manual \\
    GCN~\cite{kipf2016semi}                & 16   & $0.50$      & $71.89\pm0.33$        & manual            & $0.50$        & $68.50\pm0.98$          & manual \\
    GIN~\cite{xu2018powerful}              & 4    & $0.10$      & $85.59\pm0.01$        & manual            & $0.10$        & $58.38\pm0.24$          & manual \\
    GIN~\cite{xu2018powerful}              & 16   & $0.51$      & $85.38\pm0.14$        & manual            & $0.52$        & $64.72\pm1.55$          & manual \\
    GraphSage~\cite{hamilton2017inductive} & 4    & $0.10$      & $50.52\pm0.00$        & manual            & $0.10$        & $50.45\pm0.15$          & manual \\
    GraphSage~\cite{hamilton2017inductive} & 16   & $0.50$      & $50.49\pm0.01$        & manual            & $0.50$        & $63.84\pm0.11$          & manual \\
    GAT~\cite{velivckovic2017graph}        & 4    & $0.11$      & $75.82\pm1.82$        & manual            & $0.11$        & $57.73\pm0.32$          & manual \\
    GAT~\cite{velivckovic2017graph}        & 16   & $0.53$      & $78.27\pm0.18$        & manual            & $0.53$        & $70.59\pm0.45$          & manual \\
    GatedGCN~\cite{bresson2017residual}    & 4    & $0.10$      & $84.48\pm0.12$        & manual            & $0.10$        & $60.40\pm0.42$          & manual \\
    GatedGCN~\cite{bresson2017residual}    & 16   & $0.50$      & $86.51\pm0.09$        & manual            & $0.50$        & $76.08\pm0.20$          & manual \\
    GNAS-RL~\cite{gao2020graph}            & 4    & $0.48$      & $85.21\pm0.01$        & 5                 & $0.48$        & $52.61\pm0.22$          & 5      \\
    GNAS-MP~\cite{cai2021rethinking}       & 4    & $0.35$      & $86.80\pm0.10$        & 0.10              & $0.38$        & $62.21\pm0.20$          & 0.05   \\
    GNAS-MP~\cite{cai2021rethinking}       & 16   & $1.60$      & $\bm{86.85\pm0.10}$   & 0.10              & $1.61$        & $74.77\pm0.15$          & 0.10   \\
    \midrule 
    EGNAS (w/o E)                          & 4    & $0.09$      & $85.08\pm0.11$        & 0.08              & $0.17$        & $61.21\pm0.22$          & 0.05    \\
    EGNAS (E)                              & 4    & $0.17$      & $85.20\pm0.13$        & 0.08              & $0.32$        & $72.76\pm0.13$          & 0.05    \\
    EGNAS (w/o E)                          & 16   & $0.24$      & $86.15\pm0.15$        & 0.14              & $0.31$        & $71.80\pm0.25$          & 0.12    \\
    EGNAS (E)                              & 16   & $0.44$      & $86.20\pm0.11$        & 0.14              & $0.61$        & $\bm{76.65\pm0.11}$     & 0.12    \\
    \bottomrule
    \end{tabular}
    \vspace{-0.5 em}
\end{table}

\end{document}